%% file: main.tex
\documentclass{article}

\usepackage[margin=1in]{geometry}

\usepackage{graphicx}

\usepackage{amsmath,amssymb,amsfonts}
\usepackage{nicefrac}

\usepackage[utf8]{inputenc}
\usepackage[T1]{fontenc}
\usepackage{lmodern}
\usepackage{microtype}
\usepackage{booktabs}

\usepackage[hidelinks]{hyperref}
\usepackage{url}
\usepackage{doi}
\usepackage[numbers,square]{natbib}

\usepackage{caption}
\usepackage{subcaption}

\usepackage{enumitem}

\usepackage{booktabs}
\usepackage{graphicx}
\usepackage{longtable}
\usepackage{array}
\usepackage{ragged2e}
\usepackage{multirow}

\usepackage[table]{xcolor}
\usepackage{soul}

\usepackage{pdflscape}

\definecolor{PublicPink}{HTML}{EAD1DC}
\definecolor{LocalPurple}{HTML}{B4A7D6}

\hypersetup{
    pdftitle={Protecting patient privacy in clinical foundation models: Technical and legal perspectives},
    pdfauthor={Sana Tonekaboni, Lena Stempfle, Sasha Ronaghi, Corinna Coupette, I. Glenn Cohen, Emily Alsentzer, and Marzyeh Ghassemi},
    pdfkeywords={patient privacy, clinical foundation models, healthcare, privacy law, machine learning}
}

\begin{document}

\begin{flushleft}

{\LARGE\bfseries
Protecting Patient Privacy in Clinical Foundation Models:
Technical and Legal Perspectives\par
}

\vspace{1em}

Sana Tonekaboni$^{1,2,\dag}$,
Lena Stempfle$^{1,2,\dag}$,
Sasha Ronaghi$^{3,1}$,
Corinna Coupette$^{4,5,6}$,
I. Glenn Cohen$^{7,8}$,
Emily Alsentzer$^{3}$,
Marzyeh Ghassemi$^{1}$

\vspace{0.8em}

{\small
$^{1}$Massachusetts Institute of Technology (MIT), Cambridge, Massachusetts, USA\\
$^{2}$Broad Institute of MIT and Harvard, Cambridge, Massachusetts, USA\\
$^{3}$Stanford University, Stanford, California, USA\\
$^{4}$Aalto University, Espoo, Finland\\
$^{5}$Max Planck Institute for Tax Law and Public Finance, Munich, Germany\\
$^{6}$Stanford Center for Legal Informatics, Stanford Law School, Stanford, California, USA\\
$^{7}$Harvard Law School, Cambridge, Massachusetts, USA\\
$^{8}$Petrie-Flom Center for Health Law Policy, Biotechnology, and Bioethics,
Cambridge, Massachusetts, USA\\
$^{\dag}$Equal contribution
}



\end{flushleft}

\vspace{1.5em}

\begin{abstract}
	Clinical foundation models trained on large-scale patient data are increasingly used for decision support, screening, and public health planning. As deployment expands, privacy risk arises from model-mediated leakage, yet its prevalence and severity remain poorly quantified. Models can disclose sensitive training artifacts, enabling patient re-identification in ways not captured by data-handling controls alone. As a result, existing frameworks, including HIPAA and GDPR, offer limited protection against assessing and addressing. We propose a practical framework for assessing privacy risk in clinical foundation models, illustrate realistic leakage scenarios across deployment settings, map them to legal regimes, and outline complementary technical and legal mitigations. Our analysis provides a context-aware risk assessment grounded in realistic usage to preserve the value of medical foundation models while rigorously safeguarding patient privacy. 
\end{abstract}


\section{Introduction}
\raggedbottom
Clinical foundation models (FMs) are rapidly integrating into clinical decision-making, triage, screening, and public-health planning. Pretrained on large patient cohorts, they can be adapted to data-scarce settings and used to support diagnosis and personalized risk prediction and to streamline clinical workflows with the potential to improve patient outcomes~\citep{he2025foundation}. 
As FMs grow in scale and are adapted more broadly, privacy risk can arise from unintended data leakage, enabling patient re-identification even when trained on records where personal identifiers have been removed from the data~\citep{benitez2010reidentification}. 

\begin{figure*}[!htbp]
    \centering

    \begin{subfigure}[t]{0.85\textwidth}
        \centering
        \includegraphics[width=\linewidth]{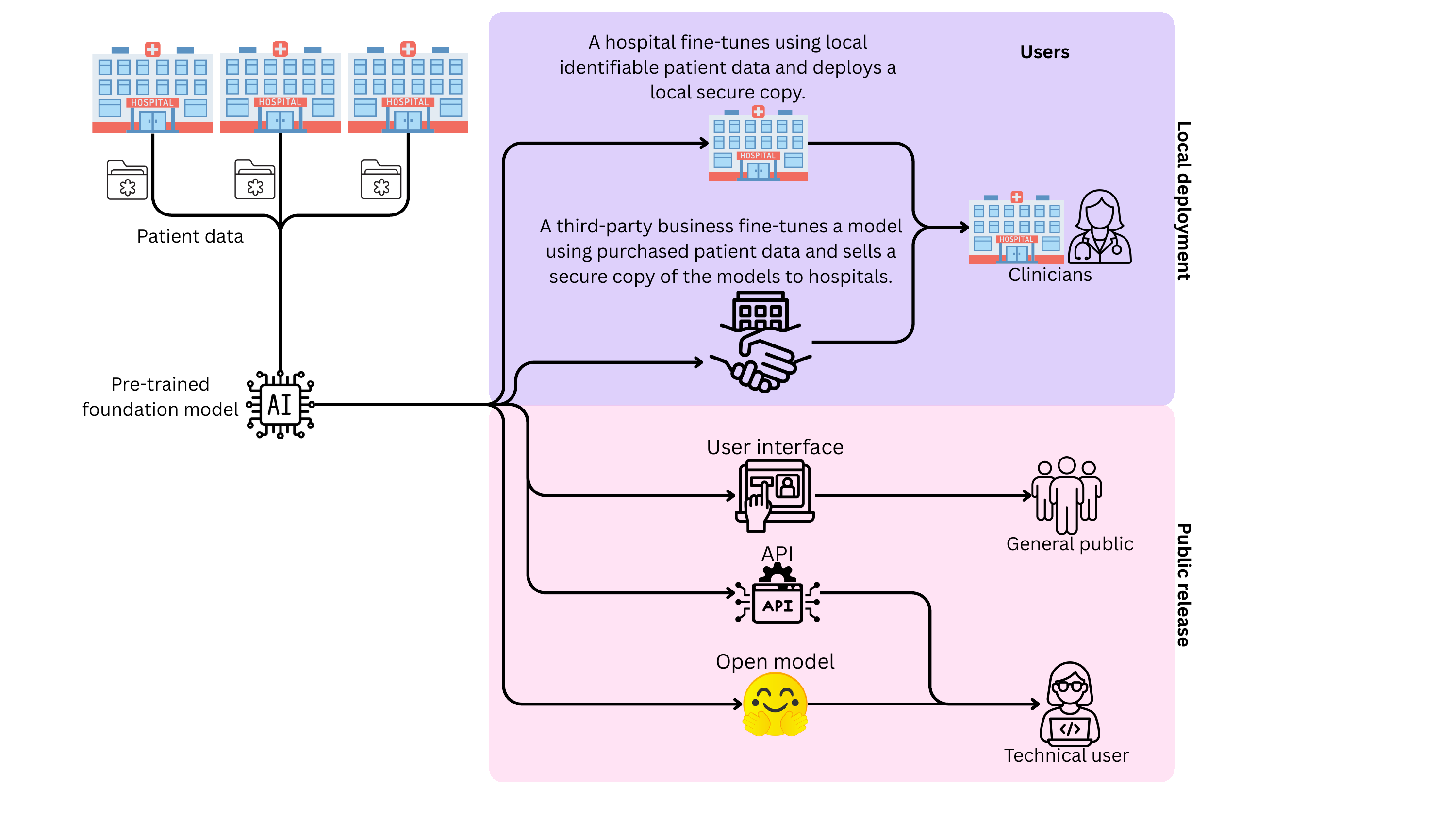}
        \caption{\textbf{Foundation model development and release workflow.} Patient data processed to reduce identifiability from multiple institutions (e.g., hospitals) are used to pre-train a foundation model (left). For this model, two deployment pathways are shown: (i) Local deployment, where the foundation model is fine-tuned by a healthcare institution or a third-party business to be securely deployed within a hospital, or (ii) public release of the model as open weights or an interface for general users. All stakeholders are assumed to operate within applicable legal frameworks, such as HIPAA, GDPR, and the EU AI Act.}
        \label{fig:workflow}
    \end{subfigure}
    \hfill
    \begin{subfigure}[t]{0.83\textwidth}
        \centering
        \includegraphics[width=\linewidth]{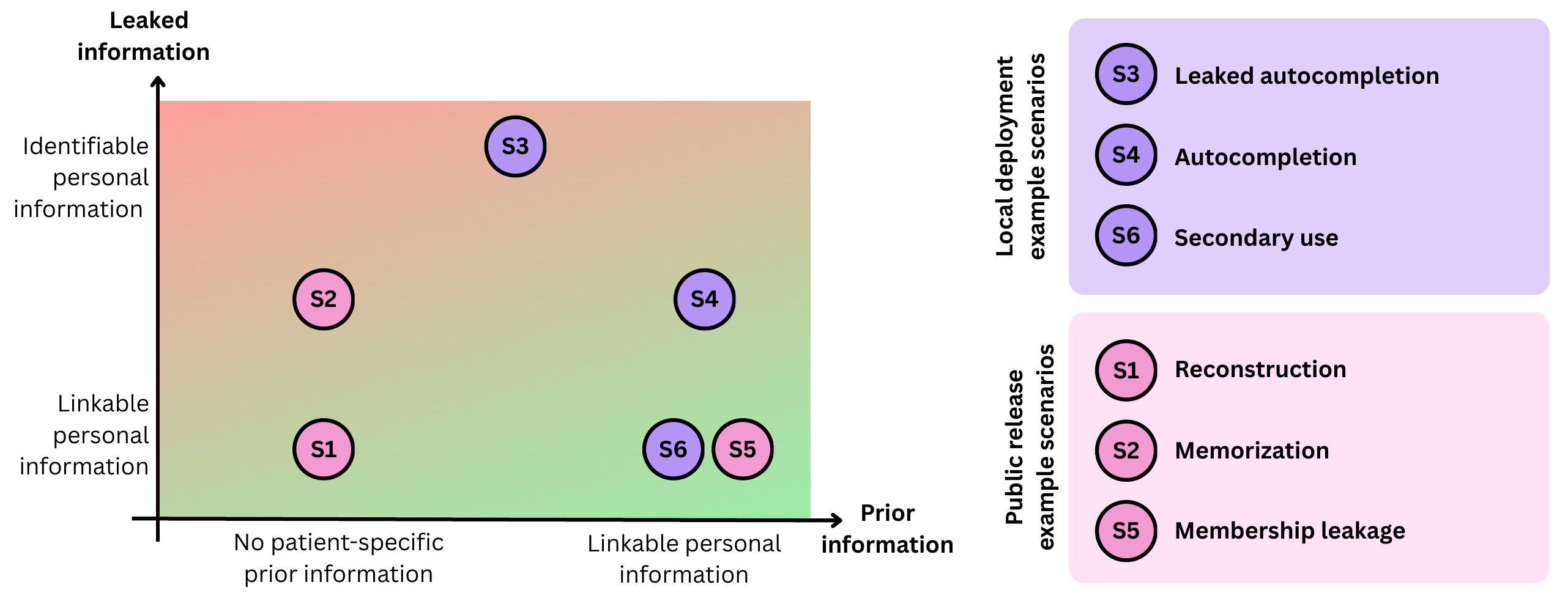}
        \caption{\textbf{Two-dimensional privacy risk framework for model-based data leakage.} The framework characterizes the privacy risk of data leakage along two orthogonal axes. Axis 1 (x-axis: Prior information) captures what information is required to extract information from a model, ranging from no patient-specific prior information (e.g., general demographic or population-level information) to information linked to a particular individual (e.g., medications or diagnoses). Axis 2 (y-axis: Leaked information) captures what type of information can be extracted from the model, progressing from linkable personal information to identifiable personal information. \textcolor{black}{We position different real-world scenarios (described in Table \ref{tab:privacy_scenarios}) within this framework.}}
        \label{fig:risk_framework}
    \end{subfigure}
    \vspace{1em}
        \centering
    \caption{Overview of the deployment workflow and privacy risk framework.}
    \label{fig:overview}
\end{figure*}

Most clinical FMs are trained on data that have been processed to reduce identifiability, rather than on fully identifiable patient records and deployed through two main pathways (Figure~\ref{fig:overview}). 
First, (i) \textbf{Local deployment}, where locally deployed copies of the FMs are hosted within a health system’s secure environment. 
Hospitals may fine-tune models on local identifiable data, while third-party vendors may fine-tune models for integration into hospital workflows. In both cases, deployment may remain within controlled clinical environments.

The second distribution type is (ii) \textbf{Public release}, in which the model is distributed as a web application with a user interface, exposed as a ``black-box'' service via an application programming interface (API), or released as downloadable weights that can be used as-is or further fine-tuned. These release channels differ materially in who can interrogate the model and under what controls, which in turn shapes the privacy threat surface.

Foundation models trained on personal health data can reveal sensitive information through their outputs, even when the original health records are not directly accessible. In this study, we focus on the privacy risk of information leakage that stems from model fragility and weakness rather than from explicit attacks such as social engineering, insider misuse, or targeted model exploitation. \textcolor{black}{Such leakage can occur unintentionally through routine interactions and emerge from fundamental model properties, not only from sophisticated or malicious attack methods.} These risks are also orthogonal to familiar clinical-environment threats and large-scale data breaches \citep{hhs2025breachreport}, which in practice often provide a simpler path to obtain sensitive information than model-centric attacks. Nevertheless, leakage through model behavior is not well addressed by existing regulatory frameworks such as HIPAA (Health Insurance Portability and Accountability Act)~\citep{hhs1996hipaa} and the GDPR (General Data Protection Regulation)~\citep{eu2016gdpr},
because existing frameworks were primarily designed to govern static risks to the direct handling of data (e.g., data collection, access, and disclosure). 
As a result, they provide limited guidance for assessing and mitigating context-dependent, model-mediated leakage that can arise through learned parameters and generated outputs.

\textcolor{black}{In this work, we examine leakage mechanisms in clinical FMs, propose a practical framework that characterizes privacy risk by (i) the prior information required to trigger leakage and (ii) the type and extent of information revealed, situate these risks within current legal regimes, and outline complementary technical and legal mitigation strategies. Through realistic scenarios, we show how the framework can distinguish technically possible leakage from plausible disclosure across deployment settings.}

\section{Contextualized Privacy Risk Assessment of Foundation Models}

To better capture and understand the contextual privacy risk of clinical foundation models, we characterize information leakage along two orthogonal axes (Figure~\ref{fig:risk_framework}): (i) the \textbf{prior information} that initiates a leak, (ii) the \textbf{type and sensitivity} of the information that is leaked by a model. 
In a practical analysis, privacy risk is determined jointly by what the model can leak and how likely it is that the required prior information is available or can be obtained in the relevant setting.

\subsection*{Axis I - Prior information}
The first axis captures the minimum amount and type of patient-specific information required to elicit or verify leakage. It ranges from attacks that require no information about a particular person and proceed through broad or unconditional model probing to attacks that begin with information linked to a specific individual. \textcolor{black}{Public availability does not determine where information falls along this axis: public records, social-media posts, and location traces may themselves be highly distinctive and patient-specific. The position of a scenario, therefore, depends on the information required to initiate the leakage, together with the auxiliary information and technical capabilities available to the potential adversary.}

\begin{itemize}[leftmargin=*]
    \item \textit{No patient-specific prior information:}
    The user does not begin with information linked to a particular person, although they may draw on general publicly available or population-level knowledge. A central scenario in this tier is unconditional generation, in which the model is probed broadly, without targeting a known patient, to elicit memorized training content, sensitive information about an arbitrary individual, or population-level information. Models may memorize and reproduce training examples, including medical images, while carefully crafted prompts can sometimes extract snippets of real data or highly specific clinical patterns even from models trained on data described as anonymized~\citep{dar2026memorize}.

    \item \textit{Linkable personal information:} 
    \textcolor{black}{Information linked to a particular person. This may include publicly accessible information, such as public records, social media posts, location traces, or visible health characteristics, as well as non-public information, such as a medication list, diagnostic test, medical image, or referral note. 
    } In such cases, the privacy risk arises from the model’s ability to infer additional sensitive details from these partial inputs, instead of information already held by the user. Generative and multimodal foundation models can use sparse cues, such as a medical image or partial medication regimen, and elicit further details such as comorbidities, hospitalization patterns, or rare diagnoses. 
    
\end{itemize}

\subsection*{Axis II - Leaked information}
Along the second axis, we categorize what type of information can be extracted from a clinical foundation model. This distinction matters because different leakage modes translate into different clinical and social harms. Some disclosures are largely reputational or administrative, while others can expose diagnoses, treatments, or highly identifying clinical narratives specific to an individual. 
We group information leakage into two  categories:
\begin{itemize}[leftmargin=*]
    \item \textit{Linkable personal information:} This category is similar to the definition in Axis I. Here, however, the category refers to information disclosed by the model rather than information already available to the user. Generative models trained on data treated as de-identified under HIPAA by stripping the required identifiers or anonymous under the GDPR may nevertheless memorize and reproduce training examples~\citep{dar2026memorize}. Such outputs may subsequently become identifying when combined with auxiliary information. However, leakage of an isolated data element that lacks direct identifiers may pose a lower immediate risk, e.g., a generative model producing a replica of a patient’s medical image released without direct identifiers. Risk arises when the output can be linked with auxiliary information, where de-identified data may be combined with (quasi-)identifiers or external data to increase privacy risk \textcolor{black}{\citep{lermen2026deanonymization}}. 
    
    \item \textit{Identifiable personal information:} This includes any personal information that could render an individual identifiable, including data elements commonly removed during de-identification (name, address, etc.). For instance, generative models introduce an additional leakage pathway by reproducing memorized fragments of private training data, either verbatim or in near-verbatim form. 
\end{itemize}

This two-axis framework helps us locate data leakage in a practical, contextual risk space.
\textcolor{black}{
Leakage risk in generative models hinges on membership inference success. Generative probing can produce candidate records, but plausibility alone does not establish training data origin. Verification requires membership inference, similarity-based ranking, or comparison with known records. Importantly, membership inference can itself cause privacy harm. For example, determining that someone was included in a cohort of patients with a particular condition may reveal that the person has—or was evaluated for—that sensitive condition. The severity of such harm depends on what incremental information the model reveals beyond an adversary's existing knowledge and which actors could plausibly acquire the requisite prior information. Moreover, recent findings show that LLMs can generate verbatim or near-verbatim content even when that content was not in their training data. Researchers evaluating privacy leakage must therefore compare model outputs with the training corpus to determine whether the model has genuinely reproduced training data or generated similar content by coincidence.}


\textcolor{black}{We operationalize the framework by specifying the deployment setting and information plausibly available to the user, then assigning each scenario according to the minimum patient-specific information required to elicit leakage (Axis I) and the most sensitive information plausibly disclosed (Axis II).}  \textcolor{black}{These qualitative positions may change with user access, auxiliary information, or deployment conditions and can inform testing, access, monitoring, and mitigation decisions.}



\section{Legal interpretation of model-mediated leakage}

Whether the described leakages trigger legal obligations depends on whether the revealed information qualifies as personal data under applicable law. Legal definitions of personal data distinguish between information whose processing is subject to data protection law and information that falls outside the scope of that law. Central to this distinction is whether an individual can be considered identifiable from the data, which determines how their processing is covered by the law~\citep{cobbe2026relative}.
We therefore examine how these leakage mechanisms intersect with existing legal frameworks in the United States and the European Union. 

In the United States, HIPAA applies to \emph{protected health information} (PHI) handled by covered entities and their business associates. It provides two methods for establishing that health information has been \emph{de-identified}. Under the Expert Determination method, a qualified expert must determine and document that the risk of identification is very small. Under the more commonly used Safe Harbor method, a covered entity must remove 18 categories of identifiers and have no actual knowledge that the remaining information could identify an individual, alone or in combination with other information (45 C.F.R. § 164.514(b)). Evidence that a model reproduces patient information or enables records to be linked to individuals could therefore call into question whether the actual-knowledge requirement is satisfied, although how this requirement applies to model-mediated leakage remains legally uncertain.

In the EU, the GDPR applies to all personal data concerning data subjects located in the EU. For data to qualify as personal data under the GDPR, it suffices that an individual remains identifiable (Article 4(1)), i.e., that information can reasonably be linked to a specific individual, considering the means available to the relevant recipient. \emph{Data concerning health} form a special category (Article 9(1)) whose processing is subject to stricter requirements, and only \emph{anonymous} data lie outside the scope of the Regulation. As a result, \emph{pseudonymized} data generally fall under the GDPR, and model outputs may constitute personal data even without direct identifiers. Training and deployment must therefore satisfy requirements including lawful basis, purpose limitation, transparency, data minimization, security, and, where applicable, erasure. The precise responsibilities depend on whether hospitals and developers act as controllers, joint controllers, or processors. 

The EU AI Act adds risk-management obligations tied to application-based risk categories, especially for providers and deployers of models, but it does not establish privacy thresholds for memorization, reconstruction, or membership inference. Clinical AI is classified as high-risk when it is a safety component of, or itself constitutes, a regulated medical device requiring third-party conformity assessment. Contracts, data-use agreements, and institutional oversight may provide additional safeguards but do not replace statutory obligations. In the future, the processing of personal health data will additionally be covered by sectoral regulation (e.g., the European Health Data Space Regulation is currently in its transition period).

\textcolor{black}{Notably, while the US and EU legal frameworks differ in their scope as well as in the relevance and interpretation of key terms (e.g., de-identification, pseudonymization, and anonymization), they provide limited guidance on how privacy safeguards should be calibrated to the context-dependent risks of model-mediated leakage.
By demonstrating how realized privacy risk depends jointly on the prior information required and the information revealed, our framework motivates a more context-sensitive approach to operationalizing these protections, in which the intensity of testing, oversight, and safeguards is calibrated to the risk presented by a particular model and deployment context rather than prescribed through fixed technical requirements.} 

\section{Risk analysis case studies}
\begin{figure}[!ht]
    \centering
    \includegraphics[width=1\textwidth]{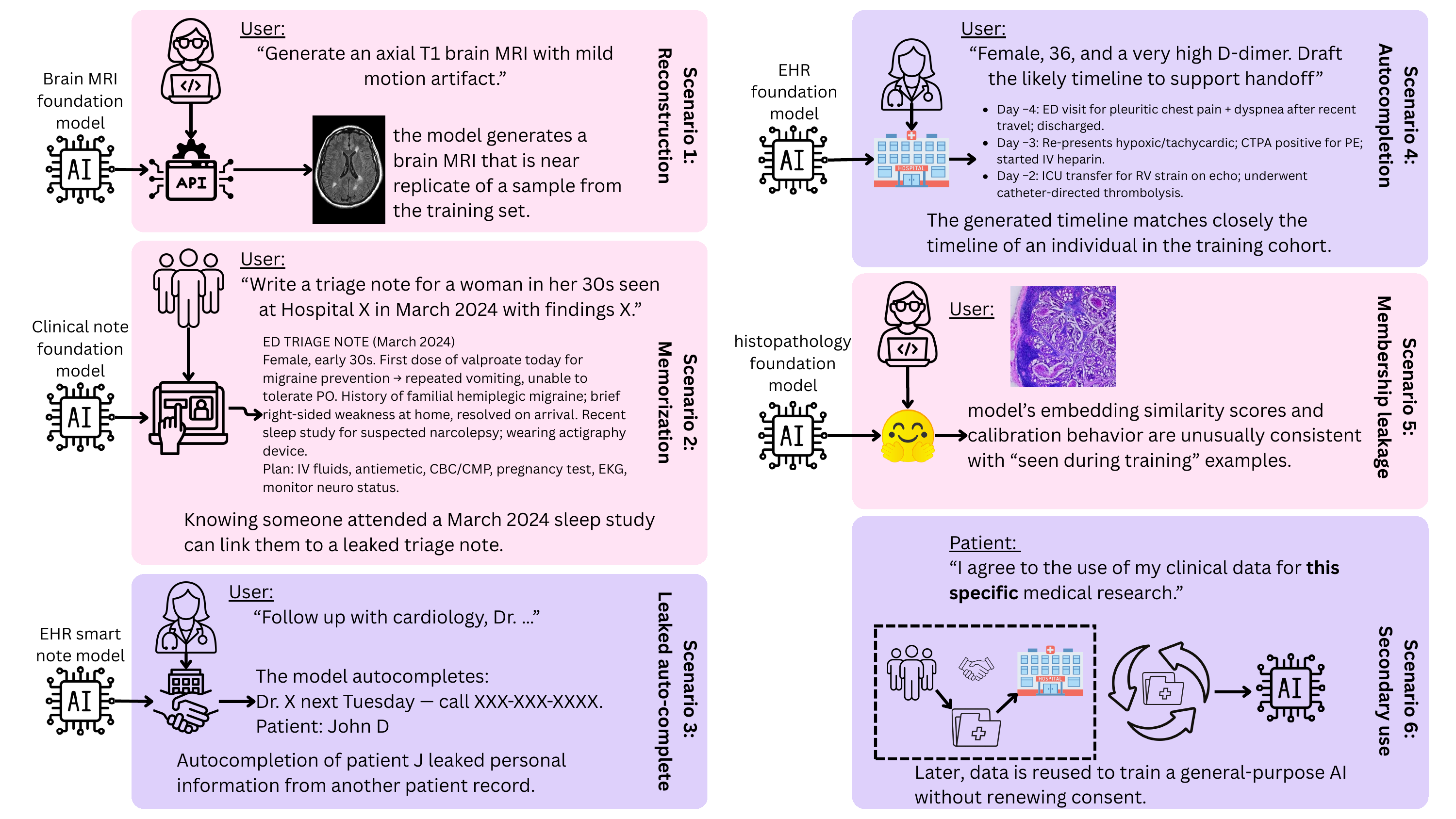}
    \caption{Privacy leakage scenarios across deployment pathways. Illustrative examples of the six scenarios described in Table~\ref{tab:privacy_scenarios}, spanning common deployment pathways for healthcare foundation models. Panels are grouped by deployment setting (pink: public release; purple: local/hospital deployment) and highlight distinct release/deployment routes within each setting, including releasing an API (S1), releasing a web interface (S2), deploying a third-party model within a hospital environment (S3), using a locally fine-tuned hospital model (S4), and public release of model weights (S5). Data originally consented for clinical research are later reused to train a general-purpose AI system without renewed consent (S6). The layout groups scenarios by deployment pathway and does not imply a risk ranking; their positions in the two-axis framework are shown in Figure 1b.}
    \label{fig:privacy_leakage}
\end{figure}

In this section, we present various realistic leakage scenarios and analyze their risks along our two-axis framework. These scenarios are summarized in Table~\ref{tab:privacy_scenarios} and Figure~\ref{fig:privacy_leakage}, and are positioned within our two-axis risk space in Figure~\ref{fig:risk_framework}.

Each row in Table~\ref{tab:privacy_scenarios} presents an example scenario of data leakage that corresponds to one of the deployment pathways depicted in Figure~\ref{fig:overview}, and is color-coded by Local deployment (purple) or Public release (pink). The ``intended use'' defines the purpose, which shapes regulatory obligations. The ``training data'' clarifies what kinds of data were used, affecting whether personal or special-category data are involved. The ``privacy risk analysis'' details the risk associated with data leakage as a function of the prior information. Finally, ``GDPR/AI Act risk analysis'' sketches legal considerations under the GDPR and the AI Act, including the relevant roles for each specific scenario. 

\textcolor{black}{Taken together, these scenarios expose a limitation of current legal frameworks: leakage events with substantially different risk profiles may receive similar legal treatment once the disclosed information falls within a protected category. The comparison in the table below, therefore, illustrates that legal classification alone may not fully capture how privacy risk varies with both the prior information required to trigger leakage and the information ultimately revealed.}


\clearpage          
\input{Tables/privacy_scenarios_two_pages_landscape_fullwidth}
\clearpage    


\section{Implications and Mitigation Strategies for Privacy Risks}
Privacy risks in clinical FMs arise across the full lifecycle, from data collection and curation to model development and deployment. They therefore cannot be addressed through a single safeguard at one point in the pipeline. \textcolor{black}{The two-axis framework translates evidence of model leakage into practical decision making. Leakage requiring no patient-specific information calls for broader testing and stronger access controls than leakage requiring information available only in a restricted clinical setting. Similarly, the disclosure of identifiable or highly sensitive information warrants stronger monitoring and remediation than the disclosure of less sensitive information. The appropriate response, therefore, depends on both axes and on the deployment setting.} Mitigation should thus be calibrated both to the lifecycle stage and to a contextual risk space. Below, we discuss these mitigation strategies at the various levels of the model life cycle. 

\paragraph{Data collection}
Data protection begins with lawful data use, minimization, and effective de-identification or pseudonymization~\citep{morley2022governing}. These safeguards must be applied consistently across contributing institutions, as differences in de-identification practices can produce uneven protection and enable source inference. 
\textcolor{black}{Rare and low-frequency records require particular attention because they may be disproportionately vulnerable to memorization and extraction~\citep{kulynych2022disparate}. Larger or more diverse datasets may reduce repeated exposure to some records, but they do not eliminate memorization, which has also been demonstrated in models trained on large, heterogeneous corpora~\citep{carlini2023quantifying}.
Dataset diversity should therefore not be treated as a privacy safeguard without direct evaluation of record-level leakage.}

\paragraph{Model development}
Modern privacy frameworks and guidance emphasize risk management and data protection by design, but offer limited guidance on privacy harms arising from model behavior rather than explicit data disclosure, creating a compliance gap.
In practice, developers should evaluate model-mediated leakage rather than relying only on the de-identification of training data. Privacy audits can include membership-inference tests, targeted extraction, canary exposure, and structured red-teaming. \textcolor{black}{Training-time design choices can also help reduce memorization risk.}
\textcolor{black}{For example, increasing model size can increase memorization risk \citep{carlini2023quantifying}; practitioners should therefore assess model and dataset scaling jointly and combine scaling decisions with strategies such as reducing repeated exposure and carefully ordering training data. Data ordering can also matter, as examples encountered earlier in training appear less likely to be memorized than those seen later~\citep{wei2026hubble}. Differential privacy can provide formal protection against the influence of individual training records~\citep{dwork2014algorithmic}. By contrast, output filtering and safety fine-tuning may block known disclosure patterns but do not remove memorized information from model parameters and may be circumvented through repeated or adaptive querying. They should therefore be treated as supplementary controls rather than privacy guarantees. Targeted retraining may address known leakage, but all mitigations should be evaluated against realistic attacks and deployment conditions.}

\paragraph{Model deployment or release}
Deployment conditions determine who can access a model, what prior information they can use, and how extensively they can probe it. Controls should therefore reflect the model’s position in the two-axis risk space. \textcolor{black}{Models capable of revealing information that directly identifies a patient or can reasonably be linked to a specific patient from limited or publicly available inputs may require restricted access, rate limiting, logging, monitoring for extraction-like behavior, and output filtering as a secondary safeguard.} Unrestricted APIs, downloadable weights, and open fine-tuning interfaces create greater risk because they permit repeated probing, modification, and redistribution beyond the original governance context.
Privacy evaluation should continue after release because model behavior and deployment contexts may change. Targeted extraction tests and memorization audits can detect newly identified leakage pathways~\citep{ICLR2025_e76814a0}. Where leakage is found, developers should restrict access and apply appropriate remediation, including filtering, targeted retraining, or machine unlearning where feasible~\citep{li2026llm}. Clear, testable criteria are still needed to determine when detected leakage requires corrective action.

\section{Conclusion} 
Clinical foundation models offer substantial potential for clinical decision support, screening, and public health, but their safe deployment requires privacy evaluation that goes beyond controlling inputs and access. Privacy risk in these systems is continuous rather than categorical. In this work, we introduce a structured, two-dimensional framework for assessing model-mediated privacy risk by jointly characterizing (i) prior information required to elicit leakage and (ii) the type and extent of information that can be revealed. This framing provides a practical lens for comparing risk across deployment settings, identifying plausible leakage pathways, and mapping them to appropriate technical safeguards and governance requirements. 

Our analysis also shows that privacy harm cannot be determined by identifiability alone. It depends on the sensitivity of the information revealed, the ease with which it can be linked to an individual, the information available to the recipient, and the context in which the model is accessed. \textcolor{black}{The central legal gap is therefore not the absence of different categories of protected information, but the difficulty of applying legal rules that center on pre-specified categories of information to privacy risks that are contextual, technically complex, and continually evolving.} Privacy evaluations should therefore connect measurable model leakage with realistic prior information, potential patient harm, and safeguards across data collection, model development, and deployment. Such context-aware evaluation is necessary to preserve the clinical value of foundation models while providing meaningful protection for patient privacy.

\newpage
\section*{Acknowledgments}
\paragraph{Funding:}
The Eric and Wendy Schmidt Center at the Broad Institute of MIT and Harvard, Postdoctoral fellowship (ST).
Wallenberg Foundation Scholarship Program for Postdoctoral studies at Massachusetts Institute of Technology and the Broad Institute (LS). 
Novo Nordisk Foundation Grant for a scientifically independent International Collaborative Bioscience Innovation \& Law Programme (Inter-CeBIL Programme, grant number NNF23SA0087056) (IGC).
The Weill Cancer Hub West Initiative (SR, EA). 
ELLIS Institute Finland (CC). 
National Science Foundation (NSF) 22-586 Faculty Early Career Development Award (no.2339381) (MG).
The AI2050 Program at Schmidt Sciences (MG).

\paragraph{Author contributions:}
Conceptualization: ST, LS, MG
Supervision: MG, CC, EA, IGC
Methodology: ST, LS, SR, CC, IGC, EA, MG
Visualization: ST, LS
Writing - original draft: ST, LS
Writing - review \& editing: ST, LS, SR, CC, IGC, EA, MG

\paragraph{Competing interests:}
IGC is a member of the Bayer Bioethics Council, a bioethics advisor for Bexorg, and an advisor for World Class Health and Manhattan Neuroscience LLP. He recently concluded service as the chair of the ethics advisory board for Illumina. He was also compensated for participating in a roundtable by Generation Patient, speaking at events organized by Philips with the Washington Post as well as the Doctors Company, attending the Transformational Therapeutics Leadership Forum organized by Galen Atlantica, and retained as an expert in health privacy, gender-affirming care, and reproductive technology lawsuits.

\paragraph{Data, code, and materials availability:}
No custom code, datasets, or unique materials were generated for this study. All information supporting the conclusions is contained in the manuscript.

\bibliographystyle{unsrtnat}
\bibliography{references}  







\end{document}

%% file: Tables/privacy_scenarios_two_pages_landscape_fullwidth.tex
%
%

\newcommand{\ScenarioTableHeader}{%
  \hline
  \multicolumn{2}{|l|}{\textbf{Example scenarios}}\\
  \hline
}

\newcommand{\ScenarioRight}[5]{%
  \textbf{Intended use}\par
  #1\par\vspace{0.65em}

  \textbf{Training data}\par
  #2\par\vspace{0.65em}

  \textbf{Privacy risk analysis}\par
  #3\par\vspace{0.65em}

  \textbf{GDPR/AI Act risk analysis}\par
  #4\par\vspace{0.65em}
}

\newcommand{\ScenarioStart}{\rule{0pt}{1.2em}}
\newcommand{\ScenarioEnd}{\par\vspace{0.1em}}

{
\setlength{\LTleft}{0pt}
\setlength{\LTright}{0pt}
\setlength{\tabcolsep}{6pt}      
\renewcommand{\arraystretch}{1.12}
\footnotesize

\begin{landscape}
\begin{longtable}{
>{\RaggedRight\arraybackslash}p{0.37\linewidth}
>{\RaggedRight\arraybackslash}p{0.59\linewidth}}

\caption{Example scenarios of data leakage across various deployment pathways.}
\label{tab:privacy_scenarios}\\
\ScenarioTableHeader
\endfirsthead

\ScenarioTableHeader
\endhead

\hline
\endfoot

\cellcolor{PublicPink}
\ScenarioStart
\textbf{S1 - Reconstruction:}
A company releases a brain-MRI generator model (as an API or open weights) that takes a short prompt like: “Generate an axial T1 brain MRI with mild motion artifact.” Training data included de-identified MRIs from EU hospitals or EU patients. During routine use, a user requests a handful of images with similarly generic prompts (no patient name, age, or identifiers). One generated image is a near-replica of a scan from an individual represented in the training set. The replica preserves distinctive anatomy and pathological patterns that could enable re-identification.
\par\medskip
\textcolor{blue}{\textbf{Prior information:} No patient-specific prior information (general prompt)}

\textcolor{blue}{\textbf{Leaked information:} linkable personal information (MRI image)}
&
\ScenarioRight
{Image generation for research or educational purposes, not to reproduce real patient scans.}
{Brain MRI images processed under institutional de-identification procedures.}
{A single leaked MRI contains limited information in isolation. Risk increases if the image can be linked to auxiliary information about the training cohort (e.g., that the model was trained on MRIs from Hospital X’s tumor clinic), and especially if membership inference confirms that a specific person’s scan was included.}
{The entity that trains and releases the model acts as the \textbf{controller} for the training data and must ensure lawful processing and effective safeguards against memorization. If a separate provider deploys the model via API, it may qualify as an additional \textbf{controller} for outputs. Downstream users are typically recipients unless they further process or redistribute data. 

\textcolor{black}{Generative models must document training sources and mitigate memorization risk but the AI Act does not define when outputs become personal data.}}



\ScenarioEnd
\\
\midrule

\cellcolor{PublicPink}
\ScenarioStart
\textbf{S2 - Memorization:}
A company offers a public user interface for clinical note generation. A user provides a prompt like: “Write an example triage note for a woman in her 30s seen at Hospital X in March 2024 with findings X/Y/Z.” The patient has posted online clues (a sleep-lab photo, hospital location, distinctive symptoms). The model returns text closely matching a real training note, releasing sensitive information that the person from the training cohort never shared publicly.
\par\medskip
\textcolor{blue}{\textbf{Prior information:} No patient-specific prior information (no patient-specific information)}

\textcolor{blue}{\textbf{Leaked information:} Linkable personal information (detailed patient information but without identifier)}

&
\ScenarioRight
{Clinical note drafting support, not reproducing real patient histories.}
{Clinical notes and triage records with direct identifiers removed.}
{A single generated note may not contain direct identifiers, yet risk arises when outputs can be linked to an external context. For example, combining model output with publicly available clues (such as a patient posting about participation in a sleep study) may enable attribution and reveal additional sensitive information, including training-set membership. Near-verbatim reproduction is consistent with documented evidence that LLMs can memorize and reproduce training data under certain prompts.}
{The company training and offering the API acts as \textbf{controller} for training and deployment and must ensure that measures intended to reduce identifiability, together with other technical and organizational safeguards, are effective. Contractual or research-use permissions do not override GDPR obligations if outputs are personal data.

\textcolor{black}{
The legal implications depend on whether the output becomes identifiable in context.}}



\ScenarioEnd
\\
\midrule

\cellcolor{LocalPurple}
\ScenarioStart
\textbf{S3 - Leaked autocompletion:}
A hospital purchases an EHR “smart note” model from a third-party company. Clinicians use the model to draft patient summaries. The model was fine-tuned on past discharge summaries that still contained identifiers. While writing Patient A’s discharge note, the clinician types: “Follow up with cardiology, Dr. …” and accepts an autocomplete suggestion. The model inserts a line copied from a different training example: “Follow up with Dr. X next Tuesday — call 416-XXX-XXXX. Patient: John D.” which belongs to Patient B. That PHI is now mistakenly placed into Patient A’s chart.
\par\medskip

\textcolor{blue}{\textbf{Prior information:}  Some linkable personal information (Cardiology follow up)}

\textcolor{blue}{\textbf{Leaked information:} Identifiable personal information (Patient and doctor names)}

&
\ScenarioRight
{The LLM supports discharge summary drafting for patients.}
{De-identified patient notes and identifiable discharge summaries used for fine-tuning.}
{The incident covers a disclosure of PHI as the model reproduces identifiers from training data. This leakage stems from model behaviors under normal use, rather than unauthorized access.}
{The hospital, as the data \textbf{controller} and model deployer, remains responsible for preventing such disclosures through appropriate safeguards, regardless of whether fine-tuning was internal or outsourced. Vendors involved in model training or hosting share \textbf{processor} or \textbf{joint controller} obligations.

\textcolor{black}{
High-risk clinical systems require risk management and human oversight but do not specifically address the context-dependent nature of privacy leakage.}}



\ScenarioEnd
\\
\midrule

\cellcolor{LocalPurple}
\ScenarioStart
\textbf{S4 - Autocompletion:}
A hospital deploys an EHR foundation model as a documentation assistant within a patient chart. Given structured context including patient demographics, recent laboratory results, medications, and prior notes, the model generates a draft clinical summary in response to prompts such as, \textit{“Draft the likely recent timeline to support handoff.”} In one rare case involving a female patient with markedly elevated D-dimer levels and atypical coagulation markers, the model produces a highly specific narrative describing an uncommon diagnostic pathway and medication regimen. Although framed as a probabilistic completion, the rare combination of features and highly specific management steps may mirror a distinctive historical case from the training data.
\par\medskip
\textcolor{blue}{\textbf{Prior information:}  Linkable personal information (Patient's EHR context)}

\textcolor{blue}{\textbf{Leaked information:} Some identifiable personal information (Medication regimen of a patient in training cohort)}

&
\ScenarioRight
{Clinical tool for context completion, not retrieval of real patient histories.}
{De-identified patient notes and identifiable discharge summaries used for fine-tuning.}
{If the completion stays at the level of common clinical reasoning (e.g., “consider PE workup”), privacy risk is low. However, rare or low-frequency combinations of diagnosis steps, or medication dosing patterns, can render the output record-like and plausibly attributable to a specific individual. At that point, the risk shifts from a generic inference to a plausible reconstruction.}
{Where the hospital determines the purpose and means of deployment, it acts as \textbf{controller}. The developer may qualify as \textbf{processor} or \textbf{joint controller}, depending on its influence over training and system design. Both must ensure safeguards against reconstruction and memorization.

\textcolor{black}{
High-risk clinical systems require risk management and human oversight but do not specifically address when model outputs constitute privacy-relevant reconstruction.}}



\ScenarioEnd
\\
\midrule







\cellcolor{PublicPink}
\ScenarioStart
\textbf{S5 - Membership leakage:}
To assess how training data influences model behavior, a researcher probes a publicly released histopathology foundation model trained on a breast cancer cohort. Using a digital slide patch from a known patient from which explicit patient identifiers have been removed, they observe that the model’s embedding similarity scores and calibration behavior are unusually consistent with “seen during training” examples compared to matched controls. Even without revealing the patient’s label or reproducing any identifiable image content, the interaction can support the inference that this individual’s specimen was part of the training cohort.
\par\medskip
\textcolor{blue}{\textbf{Prior information:}  Linkable personal information (Histopathology)}

\textcolor{blue}{\textbf{Leaked information:} Linkable personal information (Membership in a breast cancer training cohort)}

&
\ScenarioRight
{Histopathology analysis or representation learning, not confirmation of training set membership.}
{Patient data processed to remove explicit patient identifiers (e.g., names and medical record numbers).}
{A membership signal does not reveal full records but indicates that a person’s data was used for model training. In health care, inclusion alone may reveal sensitive health information, especially where the cohort is condition-specific (e.g., cancer patients).}
{The model provider acts as \textbf{controller} for training and deployment decisions and must mitigate such risks. API deployers share responsibility if their system exposes signals that enable inference. Users performing tests are typically recipients unless they systematically collect or publish results.

\textcolor{black}{
The AI Act requires risk mitigation for sensitive-data memorization but does not address membership inference explicitly.}}



\ScenarioEnd
\\
\midrule

\cellcolor{LocalPurple}
\ScenarioStart
\textbf{S6 - Secondary use:}
Patients at Hospital X consented to the use of their clinical data for medical research. Years later, the hospital (or a partner institution) contributed a large corpus of historical EHR notes and imaging data to a foundation-model pretraining dataset. The resulting model is used for broad clinical and commercial applications, including deployment by third parties. Patients were never explicitly informed that their data might be used to train general-purpose AI systems or commercial foundation models. Some patients later withdraw consent, but data already processed cannot be removed, and their data has already influenced the model's weights.
\par\medskip
\textcolor{blue}{\textbf{Prior information:}  Linkable personal information (Patient identifiers or quasi-identifiers)}

\textcolor{blue}{\textbf{Leaked information:} Linkable personal information (Diagnoses, treatments, notes, or medical images)}

&
\ScenarioRight
{Original: medical research\par Later: general-purpose model development and downstream deployment.}
{Patient data originally collected for research and subsequently processed to reduce identifiability before model training.}
{This scenario concerns purpose drift rather than a direct disclosure: data are reused beyond what patients reasonably expected. Once embedded in model weights, data are difficult to remove, creating tension with erasure rights.}
{The hospital, as \textbf{controller}, must ensure that any secondary use is lawful and compatible with the original purpose. Developers may qualify as controllers or joint controllers if they define training objectives. Responsibility centers on ensuring valid legal bases, transparency, and safeguards before reuse.

\textcolor{black}{
The AI Act requires data governance and documentation but does not resolve issues related to consent scope or machine unlearning.}
}



\ScenarioEnd
\\

\end{longtable}
\end{landscape}
}